# Non-asymptotic error bounds for scaled underdamped Langevin MCMC

Tim Zajic

Raytheon IDS
Woburn, MA 01821
timothy.zajic@raytheon.com

September 17, 2019

**Abstract**. Recent works have derived non-asymptotic upper bounds for convergence of underdamped Langevin MCMC. We revisit these bound and consider introducing scaling terms in the underlying underdamped Langevin equation. In particular, we provide conditions under which an appropriate scaling allows to improve the error bounds in terms of the condition number of the underlying density of interest.

## 1. Introduction

A focus of recent research has been the consideration of non-asymptotic convergence bounds for sampling schemes based on Langevin Markov chain Monte Carlo (MCMC) approaches. These approaches have been used to sample from a variety of challenging distributions in a variety of fields which involve stochastic modeling. The underlying Langevin equation is a stochastic differential equation which arose in physics and which has been studied extensively. There exist various forms of the Langevin equation and the variant of interest here is referred to as the underdamped Langevin equation.

Given a sufficiently well-behaved probability density, it is straightforward to write down a Langevin equation for which the stationary distribution is given by this density. Thus arises the possibility of simulating sample paths of the Langevin equation which should be close, in some sense, to samples drawn from the density. Issues which need to be considered are how long to run these paths for, which requires some understanding of how fast convergence takes place for the continuous Langevin equation to the stationary distribution. Another issue regards the discretization of time needed to be able to compute an approximation to the continuous time Langevin equation via, for example, an Euler approximation. Armed with an understanding of these two issues, one can get an estimate of the number of steps needed to achieve a desired accuracy, each step consisting of an Euler approximation, or alternative discretization scheme.

In this work we extend recent non-asymptotic results for Langevin MCMC to the case of scaled, or pre-conditioned, Langevin MCMC. Recent works have focused on obtaining bounds on the rate of convergence to stationarity for both underdamped and overdamped Langevin MCMC algorithms. The bounds obtained depend upon the error tolerance, the dimension of the state vector, the condition number of a particular matrix related to the density of interest and the choice of initialization.

Unlike these earlier works, our primary focus is to mitigate the dependence upon the condition number. For example, in the recent work [1] the authors derive an explicit non-asymptotic convergence bound for an underdamped Langevin MCMC algorithm. This bound depends upon the square of the condition number of the density. In certain applications the condition number can be a very large quantity. For example, when considering sensors for which accuracy along one or more coordinates is very high, while accuracy along one or more coordinates is relatively quite low. In such cases, it is of interest to improve the bound of [1] with regards to the condition number. We provide conditions under which this quadratic dependence can be weakened.

Rather than work with the underlying Langevin equations considered in these works, we consider a modified variant. Our modified underdamped Langevin equation results from introducing a scaling to the diffusion component of the Langevin equation, with an appropriate modification to the drift component which ensures the stationary distribution remains the same. By appropriate choice of the scaling, together with an additional assumption on the density of interest, we obtain bounds using similar arguments to those of [1] but with weaker dependence upon the condition number. Such a scaling, or pre-conditioning, appears in the work of [2] in the context of overdamped Langevin MCMC. The authors of [3] revisited this work from the perspective of Riemannian geometry. In studying [2,3], the work [4] is important as it indicates some issues associated with these earlier works.

The motivation for this work comes from experimentation carried out in which the scaled overdamped Langevin MCMC was found to perform much better than the non-scaled case when the condition number of the density was large.

The outline of this paper is as follows. We first review the underdamped Langevin MCMC algorithm and the associated non-asymptotic bound appearing in [1]. We then introduced the scaled underdamped Langevin MCMC algorithm and state our bound. We follow this with the proof of the bound. As our proof follows that of [1] quite closely, we limit ourselves to indicating the necessary modifications.

Research on Langevin-based MCMC approaches is evolving quickly. Before proceeding we mention three papers of interest recently released. In [5], the authors consider the scaled underdamped Langevin algorithm and show for Gaussian distributions a choice of scaling which minimizes the asymptotic variance. In [6], the authors revisit [1] with an eye towards better selecting some of the parameters in the un-scaled underdamped Langevin equation as well as present a variation of the proof technique which provides some improved constants. In [7], the class of densities considered is expanded upon by relaxing the requirements within a ball of some radius.

## 2. Background and Algorithm

The underdamped Langevin equation of interest for this work takes the form

$$\begin{aligned} dv_t &= -\gamma v_t\, dt - u\nabla f(x_t)dt + \sqrt{2\gamma u}\, dB_t \\ dx_t &= v_t\, dt \end{aligned} \quad (1)$$

with $B$ denoting standard Brownian motion and $B$, $x$ and $v$ taking values in $\mathbb{R}^d$. Here, $\gamma$ and $u$ are real parameters and $f$ is minus the logarithm of the density of interest. Given initial distribution $p_o$ we denote by $\Phi_t$ the operator mapping $p_o$ to $p_t$, the distribution of $(x_t, v_t)$. This equation has for stationary distribution the density

$$p^*(x,v) = \frac{1}{Z}\exp(-f(x) - \frac{1}{2u}|v|_2^2)$$

where $Z$ is a generic normalization constant. The marginal density, which is of interest to us, is

$$p^*(x) = \frac{1}{Z}\exp(-f(x)).$$

In [1] a straightforward algorithm is provided to generate an approximation to the stationary distribution based upon a discretization of equation (1) along with non-asymptotic bounds on the convergence. The following assumptions on the function $f$ are made there, as well as in a number of recent works, and assumed to hold here as well.

(A1) The function $f$ is twice continuously differentiable on $\mathbb{R}^d$ and there exists a positive constant $L$ such that

$$\|\nabla f(x) - \nabla f(y)\|_2 \leq L\|x - y\|_2, \quad \forall\, x, y \in \mathbb{R}^d.$$

(A2) There exists a positive constant $m$ such that

$$f(y) \geq f(x) + \langle \nabla f(x), y - x \rangle + \frac{m}{2}\|x - y\|_2^2, \quad \forall\, x, y \in \mathbb{R}^d.$$

A distribution $p^*$ defined as above with $f$ satisfying (A1) and (A2) is often referred to as a log concave smooth distribution. The quantity $\kappa = L/m$ is referred to as the condition number.

The metric we shall utilize is the Wasserstein distance of order 2, denoted $W_2$. The expression for this metric on probability measures is given by

$$W_2(\mu,\nu) \equiv \left(\inf_{\xi \in \Gamma(\mu,\nu)} \int \|x - y\|_2^2\, d\xi(x,y)\right)^{\frac{1}{2}}.$$

Here $\Gamma(\mu,\nu)$ denotes all couplings of $\mu$ and $\nu$. That is, $\xi \in \Gamma(\mu,\nu)$ if $\xi(A \times \mathbb{R}^d) = \mu(A)$ and $\xi(\mathbb{R}^d \times A) = \nu(A)$ for all measureable sets $A$.

We recall now [1, Theorem 1]: Let $p^{(n)}$ denote the distribution obtained from Algorithm 1 of [1] after n steps, starting with initial distribution $p^{(0)}(x,v) = 1_{x=x(0)}1_{v=0}$. Let the initial distance satisfy $\|x(0) - x^*\|^2 \leq \mathcal{D}^2$, where $x^*$ is the unique minimum of $f$. Setting the step size of the discretization algorithm to be

$$\delta = \frac{\varepsilon}{104\kappa}\sqrt{\frac{1}{\frac{d}{m} + \mathcal{D}^2}},$$

and running the algorithm for $n$ steps with

$$n \geq \left(\frac{52\kappa^2}{\varepsilon}\right)\sqrt{\frac{d}{m} + \mathcal{D}^2}\log\left(\frac{24(\frac{d}{m} + \mathcal{D}^2)}{\varepsilon}\right),$$

it holds that

$$W_2(p^{(n)}, p^*) \leq \varepsilon.$$

For our setting, we consider the scaled underdamped Langevin equation

$$\begin{aligned} dv_t &= -\gamma A v_t\, dt - u\nabla f(x_t)dt + \sqrt{2\gamma u A}dB_t \\ dx_t &= v_t\, dt \end{aligned} \quad (2)$$

with $\widetilde{\Phi}_t$ and $\tilde{p}_t$ defined analogously to $\Phi_t$ and $p_t$. We remark that it is possible to generalize this expression by using a state dependent matrix $A$ [4]. However, the expression for the scaled equation will also include derivatives of the matrix $A$.

A step of the discretized version of this scaled equation takes the form

$$\begin{aligned} d\tilde{v}_t &= -\gamma A\tilde{v}_t\, dt - u\nabla f(\tilde{x}_0)dt + \sqrt{2\gamma u A}dB_t \\ d\tilde{x}_t &= \tilde{v}_t\, dt \end{aligned}$$

for the initial condition $(\tilde{x}_0, \tilde{v}_0) \sim \tilde{p}_0$. The difference between this and the non-discretized version is that the argument of $f$ is 'frozen': this makes the solution particularly straightforward to compute with.

We now state the algorithm associated with our scaled underdamped Langevin equation. We remark that the essential difference from the algorithm in [1] is the introduction of the scaling $A$.

**Algorithm**:

Inputs: Parameters $u, \gamma, A$. Step size $\delta$. Number of iterations $n$. Initial point $(x^0, 0)$. Gradient oracle $\nabla f$.

for $i = 0, 1, ..., n-1$ do

sample $(x^{i+1}, v^{i+1}) \sim Z^{i+1}(x^i, v^i)$

end

The random vector $Z^{i+1}(x^i, v^i)$ is a Gaussian distribution with mean given by

$$E[x^{i+1}] = x^i + \frac{A^{-1}}{\gamma}(1 - e^{-\gamma A \delta})v^i - \frac{uA^{-1}}{\gamma}\left(t - \frac{A^{-1}}{\gamma} + \frac{A^{-1}}{\gamma}e^{-\gamma A \delta}\right)$$

$$E[v^{i+1}] = v^i e^{-\gamma A \delta} - \frac{uA^{-1}}{\gamma}(1 - e^{-\gamma A \delta})\nabla f(x^i)$$

and covariance given by

$$E[(x^{i+1} - E[x^{i+1}])(x^{i+1} - E[x^{i+1}])^T] = \frac{2uA^{-1}}{\gamma}\left(t - \frac{A^{-1}}{2\gamma}e^{-2\gamma A \delta} + \frac{2A^{-1}}{\gamma}e^{-\gamma A \delta} - \frac{3A^{-1}}{2\gamma}\right)$$

$$E[(v^{i+1} - E[v^{i+1}])(v^{i+1} - E[v^{i+1}])^T] = u(1 - e^{-2\gamma A \delta})$$

$$E[(x^{i+1} - E[x^{i+1}])(v^{i+1} - E[v^{i+1}])^T] = 2u\left(\frac{A^{-1}}{2\gamma} + \frac{A^{-1}}{2\gamma}e^{-2\gamma A \delta} - \frac{A^{-1}}{\gamma}e^{-\gamma A \delta}\right)$$

We denote by $p^{(n)}$ the distribution of $(x^n, v^n)$. Furthermore, denote by $q^*$ and $q^{(n)}$ the distribution of $(x, x + v)$, when $(x, v)$ is distributed according to $p^*$ and $p^{(n)}$, respectively.

## 3. Theorem Statement and Proof

We show here that in case $\nabla^2 f$ does not oscillate too much, bounds analogous to those appearing in [1] continue to hold with a smaller exponent on the condition number when the matrix $A$ is properly chosen.

The proof of [1, Theorem 1] makes use of two other auxiliary theorems [1, Theorem 5, Theorem 9]. The first regards convergence to stationarity for the continuous process while the second is concerned with the error induced by discretization of the continuous process. The proofs of these auxiliary theorems in turn make use of a number of lemmas. As our results parallel and leverage those of [1], we will refer the reader to the later in those cases where the proof differs trivially.

Before stating our theorem we first introduce some notation. Denote by $m^x$ the smallest eigenvalue of $\nabla^2 f(x)$ and consider the optimization problem

$$\Theta = \min_y \max_x \|\nabla^2 f(x) - \nabla^2 f(y)\|_2 / m^y.$$

Denote the value of $y$ at which the minimum occurs by $\hat{y}$, let $\hat{m} = m^{\hat{y}}$ and set

$$A = u\nabla^2 f(\hat{y}), \quad u = \frac{2}{\hat{m}}, \quad \gamma = 1.$$

By [8, Lemma 4], it holds that $\hat{m} \geq m$ and so, letting $\hat{\kappa} = L/\hat{m}$, it holds that $\hat{\kappa} \leq \kappa$. The quantity $\nabla^2 f(\hat{y})$ is sometimes called the observed Fisher matrix at the point $\hat{y}$. In the case of a Gaussian, or normal, distribution $N(\mu, \Sigma)$ the observed Fisher matrix is a constant and equals $\Sigma^{-1}$ and hence $\Theta = 0$.

With these choices for $A$ and $\gamma$ our scaled underdamped Langevin equation reads

$$\begin{aligned} dv_t &= -u\nabla^2 f(\hat{y})v_t\, dt - u\nabla f(x_t)dt + \sqrt{2u^2\nabla^2 f(\hat{y})}dB_t \\ dx_t &= v_t\, dt \end{aligned}$$

**Theorem 1.** With the above choices of parameters, in case $\Theta \leq 1/2$, let

$$\delta = \frac{\varepsilon(1-2\Theta)}{\hat{\kappa}}\sqrt{\frac{5}{73728}}\sqrt{\frac{1}{(\frac{d}{m} + \mathcal{D}^2)}}$$

For $\varepsilon$ small enough, it suffices to pick

$$n \geq \frac{\hat{\kappa}}{\varepsilon(1-2\Theta)^2}\sqrt{\frac{18432}{5}}\sqrt{\frac{d}{m} + \mathcal{D}^2}\log\left(\frac{16(2\frac{d}{m} + \mathcal{D}^2)}{\varepsilon}\right)$$

in order to ensure that

$$W_2(p^{(n)}, p^*) \leq \varepsilon.$$

**Remark:** Before proving Theorem 1, we comment on the condition that $\varepsilon$ be "small enough". In the proof of Theorem 1, we require that $2\delta(1-2\Theta) < 1$, which holds for $\varepsilon$ sufficiently small. A further constraint on the size of $\varepsilon$ appears in the proof of Lemma 2.

**Proof:** We begin with a variant of the result appearing in [1, Theorem 5]. There, consideration is given to the roots of the following characteristic equation:

$$\det\begin{bmatrix} (\gamma - 1 - \lambda)I_{dxd} & \frac{u\mathcal{H}_t - \gamma I_{dxd}}{2} \\ \frac{u\mathcal{H}_t - \gamma I_{dxd}}{2} & (1-\lambda)I_{dxd} \end{bmatrix}$$

where

$$\mathcal{H}_t \equiv \int_0^1 \nabla^2 f(x_t + h(y_t - x_t))dh.$$

In the case of the scaled underdamped Langevin, the appropriate expression is given by

$$\det \begin{bmatrix} (\gamma A - 1 - \lambda)I_{dxd} & \dfrac{u\mathcal{H}_t - \gamma A}{2} \\ \dfrac{u\mathcal{H}_t - \gamma A}{2} & (1 - \lambda)I_{dxd} \end{bmatrix}$$

Consider the case in which $\nabla^2 f$ is constant. Recalling that $\gamma = 1$ and replacing $u\mathcal{H}_t$ by $\gamma A$, the characteristic equation of interest is

$$\det \begin{bmatrix} (A - 1 - \lambda)I_{dxd} & 0 \\ 0 & (1 - \lambda)I_{dxd} \end{bmatrix}$$

Writing

$$A = \sum \alpha_i e_i e_i^T$$

for an orthonormal basis, we can diagonalize to write our characteristic equation as

$$(\alpha_i - 1 - \lambda)(1 - \lambda) = 0, \quad i = 1, \ldots, d.$$

By our choice of u we know that $\alpha_i \geq 2$ and hence all solutions of this equation satisfy

$$\lambda_i \geq 1.$$

Consider now the general case, where we are interested in the eigenvalues, $\lambda_i^{\mathcal{A}}$, of the matrix

$$\mathcal{A} = \begin{bmatrix} (A - 1)I_{dxd} & \dfrac{u\mathcal{H}_t - AI_{dxd}}{2} \\ \dfrac{u\mathcal{H}_t - AI_{dxd}}{2} & I_{dxd} \end{bmatrix}$$

We have the bound

$$\left\| \begin{bmatrix} (A - 1)I_{dxd} & 0 \\ 0 & I_{dxd} \end{bmatrix} - \begin{bmatrix} (A - 1)I_{dxd} & \dfrac{u\mathcal{H}_t - AI_{dxd}}{2} \\ \dfrac{u\mathcal{H}_t - AI_{dxd}}{2} & I_{dxd} \end{bmatrix} \right\|_2$$

$$= \left\| \begin{bmatrix} 0 & \dfrac{u\mathcal{H}_t - u\nabla^2 f(\hat{y})}{2} \\ \dfrac{u\mathcal{H}_t - u\nabla^2 f(\hat{y})}{2} & 0 \end{bmatrix} \right\|_2 = \|u\mathcal{H}_t - u\nabla^2 f(\hat{y})\|_2 \leq 2\Theta$$

From the Bauer-Fike Theorem it holds that

$$\min_i \lambda_i^{\mathcal{A}} \geq 1 - 2\Theta.$$

As in [1], the above allows to conclude that

$$E_{(x_t, v_t, y_t, w_t) \sim \zeta_t((x_0, v_0, y_0, w_0))}[|x_t - y_t|^2 + |(x_t + v_t) - (y_t + w_t)|^2]$$

$$\leq e^{-2t(1-2\Theta)}[|x_0 - y_0|^2 + |(x_0+v_0) - (y_0 + w_0)|^2]$$

for some coupling $\zeta_t((x_0, v_0, y_0, w_0))$.

Note that in [1], the multiplier is $e^{-t/\kappa}$ rather than $e^{-2t(1-2\Theta)}$.

The implication of this result is that the inequality in [1, Lemma 8] may be replaced by

$$W_2(\Phi_t p_0, p^*) \leq 4e^{-t(1-2\Theta)} W_2(p_0, p^*).$$

We now consider the relevant variant of [1, Theorem 9] for our needs. This proof relies upon modifications to [1, Lemmas 10, 12, 13], which are presented at the end of this section.

We start by noting, by Lemma 1,

$$E[|v_s - \tilde{v}_s|_2^2] = E[|u\int_0^s e^{-\gamma A(t-s)}(\nabla f(x_r) - \nabla f(x_0))dr|_2^2]$$

$$\leq su^2 \int_0^s E[|e^{-\gamma A(t-s)}(\nabla f(x_r) - \nabla f(x_0))|_2^2]dr$$

With $\gamma = 1$ and noting that

$$|e^{-A(t-s)}|_2 = \sigma_{\max}(e^{-A(t-s)}) = \max_i(e^{-\alpha_i(t-s)}) \leq 1$$

we can write

$$E[|v_s - \tilde{v}_s|_2^2] \leq su^2 \int_0^s E[|(\nabla f(x_r) - \nabla f(x_0))|_2^2]dr$$

$$\leq su^2 L^2 \int_0^s E[|x_r - x_0|_2^2]dr$$

$$= su^2 L^2 \int_0^s E[|\int_0^r v_w dw|_2^2]dr$$

$$\leq su^2 L^2 \int_0^s r(\int_0^r E[|v_w|_2^2]dw)dr$$

$$\leq su^2 L^2 \mathcal{E}_K \int_0^s r(\int_0^r dw)dr$$

$$= \frac{s^4 u^2 L^2 \mathcal{E}_K}{3}$$

where $\mathcal{E}_K$ is given in Lemma 2. As in the proof of [1, Theorem 9], this implies

$$W_2^2(\Phi_\delta p_0, \tilde{\Phi}_\delta p_0) \leq \mathcal{E}_K \left(\frac{\delta^4 u^2 L^2}{3} + \frac{\delta^6 u^2 L^2}{15}\right).$$

With our choice for u and the assumption that $\delta \leq 1$, we have

$$W_2^2(\Phi_\delta p_0, \widetilde{\Phi}_\delta p_0) \leq \frac{8\delta^4 \mathcal{E}_K \hat{\kappa}^2}{5}.$$

Moving on with the proof of Theorem 1, the first two inequalities in the proof of [1, Theorem 1] are now replaced with

$$W_2(\Phi_\delta q^{(i)}, q^*) \leq e^{-\delta(1-2\Theta)} W_2(q^{(i)}, q^*)$$

and

$$W_2(\Phi_\delta q^{(i)}, \widetilde{\Phi}_\delta q^{(i)}) \leq 2W_2(\Phi_\delta p^{(i)}, \widetilde{\Phi}_\delta p^{(i)}) \leq \delta^2 \sqrt{\frac{32\mathcal{E}_K \hat{\kappa}^2}{5}}.$$

The first inequality improves over that of [1] by not having $\kappa$ in it, while the second is worse by having a 32 rather than an 8 and the presence of $\hat{\kappa}$ due to our choice of $u$.

The triangle inequality implies

$$W_2(q^{(i+1)}, q^*) \leq \delta^2 \sqrt{\frac{32\mathcal{E}_K \hat{\kappa}^2}{5}} + e^{-\delta(1-2\Theta)} W_2(q^{(i)}, q^*).$$

Setting $\eta = \exp(-\delta(1-2\Theta))$ and applying this last expression n times yields

$$W_2(q^{(n)}, q^*) \leq \eta^n W_2(q^{(0)}, q^*) + (1 + \eta + \cdots + \eta^{n-1}) \delta^2 \sqrt{\frac{32\mathcal{E}_K \hat{\kappa}^2}{5}}$$

which in turn implies

$$W_2(q^{(n)}, q^*) \leq 2\eta^n W_2(p^{(0)}, p^*) + \left(\frac{1}{1-\eta}\right) \delta^2 \sqrt{\frac{32\mathcal{E}_K \hat{\kappa}^2}{5}}$$

and this in turn implies

$$W_2(p^{(n)}, p^*) \leq 4\eta^n W_2(p^{(0)}, p^*) + \left(\frac{1}{1-\eta}\right) \delta^2 \sqrt{\frac{128\mathcal{E}_K \hat{\kappa}^2}{5}} \equiv T_1 + T_2.$$

The first term accounts for the convergence of the continuous time process to the invariant distribution over n steps, while the second term accounts for the error due to discretization, accumulated over the n steps. We now bound both terms by $\varepsilon/2$, taking into account that

$$1 - \eta = 1 - \exp(-\delta(1-2\Theta)) \geq \frac{\delta(1-2\Theta)}{2}$$

since $2(1-2\Theta)\delta < 1$.

In bounding $T_2$ we desire that

$$\left(\frac{1}{1-\eta}\right)\delta^2\sqrt{\frac{128\mathcal{E}_K\hat{\kappa}^2}{5}} \leq \frac{\varepsilon}{2}.$$

We have

$$\left(\frac{1}{1-\eta}\right)\delta^2\sqrt{\frac{128\mathcal{E}_K\hat{\kappa}^2}{5}} \leq \frac{2\delta}{(1-2\Theta)}\sqrt{\frac{128\mathcal{E}_K\hat{\kappa}^2}{5}}.$$

Notice this differs from [1] in that we lose $\kappa$ in our bound on $\left(\frac{1}{1-\eta}\right)$, but we gain a $\hat{\kappa}$ through our upper bound of $\mathcal{E}_K\hat{\kappa}^2$. Both of these are due to our modification of the underdamped Langevin and they cancel each other. We see our desire is met if

$$\delta \leq \frac{\varepsilon}{2}\frac{(1-2\Theta)}{2}\sqrt{\frac{5}{128\mathcal{E}_K\hat{\kappa}^2}} = \frac{\varepsilon(1-2\Theta)}{\hat{\kappa}}\sqrt{\frac{5}{73728}}\sqrt{\frac{1}{\left(\frac{d}{m}+\mathcal{D}^2\right)}}.$$

In bounding $T_1$ we desire that

$$4\eta^n W_2(p^{(0)},p^*) \leq \frac{\varepsilon}{2}$$

or, equivalently,

$$n\log\eta \leq \log\left(\frac{\varepsilon}{8W_2(p^{(0)},p^*)}\right).$$

Recalling that $\eta < 1$,

$$\begin{aligned}n &\geq \frac{1}{\log(\eta)}\log\left(\frac{\varepsilon}{8W_2(p^{(0)},p^*)}\right) \\ &= \frac{1}{\delta(1-2\Theta)}\log\left(\frac{8W_2(p^{(0)},p^*)}{\varepsilon}\right).\end{aligned}$$

Note we gained an extra $\kappa$ over [1] since our $\eta$ does not depend on $\kappa$.

Using our expression for $\delta$ and Lemma 3, we have

$$n \geq \frac{\hat{\kappa}}{\varepsilon(1-2\Theta)^2}\sqrt{\frac{18432}{5}}\sqrt{\left(\frac{d}{m}+\mathcal{D}^2\right)}\log\left(\frac{16\left(2\frac{d}{m}+\mathcal{D}^2\right)}{\varepsilon}\right).$$

∎

Finally, we turn to the modifications of the supporting lemmas, corresponding to [1, Lemmas 10, 12, 13].

**Lemma 1.** The solution of (2) is given by

$$v_t = v_0 e^{-\gamma At} - u\left(\int_0^t e^{-\gamma A(t-s)} \nabla f(x_s) ds\right) + \sqrt{2\gamma u A} \int_0^t e^{-\gamma A(t-s)} dB_s$$

$$x_t = x_0 + \int_0^t v_s \, ds$$

**Proof:** The proof follows by substitution. ∎

**Lemma 2.** For $\varepsilon$ sufficiently small

$$E_{(x,v) \sim \Phi_t p^{(i)}}[\|v\|_2^2] \leq \mathcal{E}_K$$

with

$$\mathcal{E}_K = 36\left(\frac{d}{m} + \mathcal{D}^2\right).$$

**Proof:** The proof follows from a modification of that of [1, Lemma 12]. In particular, by our choice of $u$, rather than

$$p^* \propto \exp(-(f(x) + \frac{L}{2}|v|_2^2))$$

we have

$$p^* \propto \exp(-(f(x) + \frac{\hat{m}}{4}|v|_2^2))$$

so that

$$E_{p^*}[|v|_2^2] = \frac{2d}{\hat{m}}.$$

As a result, whereas in the proof of [1, Lemma 12], the following inequality is used

$$E_{p^{(0)}}[|v|_2^2] \leq \frac{10d}{L} + \frac{16d}{m} + 16\mathcal{D}^2 \leq 26(\frac{d}{m} + \mathcal{D}^2)$$

we have the bound

$$E_{p^{(0)}}[|v|_2^2] \leq \frac{20d}{\hat{m}} + \frac{16d}{m} + 16\mathcal{D}^2 \leq 36\left(\frac{d}{m} + \mathcal{D}^2\right) = \mathcal{E}_K.$$

Next, we note that in Step 4 of the proof of [1, Lemma 12], appears the statement: Then by Theorem 9 applied for $i$ steps, we know that

$$W_2^2(q^{(i+1)}, q^*) = W_2^2(\tilde{\Phi}_\delta\, q^{(i)}, q^*) \leq W_2^2(q^{(i)}, q^*).$$

It is unclear why this statement holds, though it appears that the following is what the authors had in mind. As in the beginning of the proof of [1, Theorem 1], in the setting of that paper we have

$$W_2(q^{(i+1)}, q^*) \leq \delta^2 \sqrt{\frac{8\mathcal{E}_K}{5}} + e^{-\delta/2\kappa} W_2^2(q^{(i)}, q^*).$$

Given

$$W_2(q^{(i)}, q^*) \leq \sqrt{2d\left(\frac{1}{L} + \frac{2}{m}\right) + 4\mathcal{D}^2}$$

one seeks to show that

$$W_2(q^{(i+1)}, q^*) \leq \sqrt{2d\left(\frac{1}{L} + \frac{2}{m}\right) + 4\mathcal{D}^2}.$$

Setting $A = 2d\left(\frac{1}{L} + \frac{2}{m}\right) + 4\mathcal{D}^2$, we have

$$W_2(q^{(i+1)}, q^*) \leq \delta^2 \sqrt{\frac{8\mathcal{E}_K}{5}} + e^{-\delta/2\kappa} \sqrt{A}$$

so that it suffices that

$$\delta^2 \sqrt{\frac{8\mathcal{E}_K}{5}} + e^{-\delta/2\kappa} \sqrt{A} \leq \sqrt{A}. \tag{3}$$

This holds in case

$$\delta^2 \sqrt{\frac{8\mathcal{E}_K}{5}} + (1 - \delta/4\kappa)\sqrt{A} \leq \sqrt{A}$$

or

$$\delta \leq \frac{1}{4\kappa} \sqrt{\frac{5A}{8\mathcal{E}_K}}$$

using $e^{-\delta/2\kappa} \leq (1 - \frac{\delta}{4\kappa})$, which holds for $\delta/\kappa < 1$. In our setting, we have $A = \frac{8d}{m} + 4\mathcal{D}^2$ and, rather than (3), we require that

$$\delta^2 \sqrt{\frac{8\mathcal{E}_K \hat{\kappa}^2}{5}} + e^{-\delta(1-2\Theta)}\sqrt{A} \leq \sqrt{A}.$$

With $\varepsilon$ small enough that $2\delta(1-2\Theta)<1$, this holds in case

$$\delta^2 \sqrt{\frac{8\mathcal{E}_K \hat{\kappa}^2}{5}} + (1 - \frac{\delta(1-2\Theta)}{2})\sqrt{A} \leq \sqrt{A}.$$

or

$$\delta \sqrt{\frac{8\mathcal{E}_K \hat{\kappa}^2}{5}} \leq \frac{(1-2\Theta)}{2}\sqrt{A}.$$

Hence we require $\varepsilon$ small enough that

$$\delta \leq \frac{(1-2\Theta)}{2\hat{\kappa}}\sqrt{\frac{5A}{8\mathcal{E}_K}} = \frac{(1-2\Theta)}{2\hat{\kappa}}\sqrt{\frac{5\left(\frac{8d}{m} + 4\mathcal{D}^2\right)}{8\mathcal{E}_K}}.$$

∎

**Lemma 3.** Given our choice of parameters, we have

$$W_2^2(p^{(0)}, p^*) \leq 2(2\frac{d}{m} + \mathcal{D}^2)$$

**Proof:** Note that our choice of $u$ implies

$$E_{p^*}[|v|_2^2] = \frac{2d}{\hat{m}}$$

rather than $d/L$ as in the proof of [1, Lemma 13]. Given this, proceeding as in the proof of the latter yields

$$W_2^2(p^{(0)}, p^*) \leq 2\frac{d}{m} + 2\frac{d}{\hat{m}} + 2\mathcal{D}^2 \leq 2\left(2\frac{d}{m} + \mathcal{D}^2\right).$$

∎

# 4. Conclusion

We have demonstrated that recent non-asymptotic error bounds for underdamped Langevin MCMC can be improved upon in particular cases by appropriately scaling the terms in the underlying stochastic differential equation. In particular, the bounds are improved in terms of the condition number of the underlying density of interest. While recipes for such scalings have been put forth in both the overdamped and underdamped Langevin MCMC setting, to date we are not aware of any works showing improved non-asymptotic error upper bounds. As densities with large condition numbers arise in various application areas, such bounds are of interest to the practitioner.

## ACKNOWLEDGMENTS

The author would like to thank Fred Daum for drawing his attention to and engaging in discussions regarding Langevin MCMC techniques.